\documentclass[conference,compsoc]{IEEEtran}
\IEEEoverridecommandlockouts
\usepackage{cite}
\usepackage{amsmath,amssymb,amsfonts}
\usepackage{algorithmic}
\usepackage{graphicx}
\usepackage{textcomp}
\usepackage{xcolor}
\usepackage{booktabs}
\usepackage[hidelinks]{hyperref}
\usepackage{mathtools}
\usepackage{diagbox}
\usepackage{todonotes}
\usepackage{algorithm}
\usepackage{paralist}
\usepackage{subfigure}
\def\BibTeX{{\rm B\kern-.05em{\sc i\kern-.025em b}\kern-.08em
    T\kern-.1667em\lower.7ex\hbox{E}\kern-.125emX}}

\begin{document}

\title{IB-RAR: Information Bottleneck as Regularizer for Adversarial Robustness\\
}


\author{\IEEEauthorblockN{Xiaoyun Xu}
\IEEEauthorblockA{\textit{Radboud University}\\
the Netherlands\\
xiaoyun.xu@ru.nl}
\and
\IEEEauthorblockN{Guilherme Perin}
\IEEEauthorblockA{\textit{Leiden University}\\
the Netherlands\\
guilhermeperin7@gmail.com}
\and
\IEEEauthorblockN{Stjepan Picek}
\IEEEauthorblockA{\textit{Radboud University} \\
the Netherlands\\
stjepan.picek@ru.nl}
}

\maketitle

\pagestyle{plain}

\begin{abstract}
In this paper, we propose a novel method, IB-RAR, which uses Information Bottleneck (IB) to strengthen adversarial robustness for both adversarial training and non-adversarial-trained methods.
We first use the IB theory to build regularizers as learning objectives in the loss function.
Then, we filter out unnecessary features of intermediate representation according to their mutual information (MI) with labels,
as the network trained with IB provides easily distinguishable MI for its features.
Experimental results show that our method can be naturally combined with adversarial training and provides consistently better accuracy on new adversarial examples.
Our method improves the accuracy by an average of 3.07\% against five adversarial attacks for the VGG16 network, trained with three adversarial training benchmarks and the CIFAR-10 dataset.
In addition, our method also provides good robustness for undefended methods, such as training with cross-entropy loss only.
Finally, in the absence of adversarial training, the VGG16 network trained using our method and the CIFAR-10 dataset reaches an accuracy of 35.86\% against PGD examples, while using all layers reaches 25.61\% accuracy.
\end{abstract}


\section{Introduction}
\label{sec:introduction}
%
%

Deep learning networks are vulnerable to adversarial attacks~\cite{szegedy2013intriguing,10.1007/978-3-642-40994-3_25}.
Neural network predictions can be easily fooled by subtle adversarial perturbations, while the input remains visually imperceptible to humans.
Such perturbations can be generated by specific algorithms, such as Fast Gradient Sign Method (FGSM)~\cite{DBLP:journals/corr/GoodfellowSS14}, projected gradient descent (PGD)~\cite{https://doi.org/10.48550/arxiv.1706.06083}, and Carlini \& Wagner (CW)~\cite{7958570}. 
The main goal of these algorithms is to find as small perturbations as possible that mislead the prediction model.
This potential vulnerability raises concerns about the reliability of practical deep learning applications, 
especially in security-sensitive fields, such as vulnerability detection~\cite{9448435}, drug discovery~\cite{doi:10.1021/ci500747n}, and financial market predictions~\cite{FISCHER2018654}.


To design efficient defenses against adversarial attacks, it is also important to understand the main reasons for their success. 
Previous works have proposed many possible causes for successful adversarial attacks. 
Goodfellow et al.~\cite{DBLP:journals/corr/GoodfellowSS14} argued that the adversarial examples are generated by the excessive linearity behavior of DNNs in high-dimensional spaces.
Tanay and Griffin~\cite{https://doi.org/10.48550/arxiv.1608.07690} argued that the existence of adversarial examples depends on the dislocation of classification boundary and the sub-manifold of each class. 
Some other works also regarded the primary cause for the adversarial examples as the high dimension nature of the data manifold~\cite{https://doi.org/10.48550/arxiv.1801.02774}, and adversarial attack is inevitable for a broad class of problems~\cite{https://doi.org/10.48550/arxiv.1809.02104}.
Ilyas et al.~\cite{NEURIPS2019_e2c420d9} have demonstrated that adversarial attacks can arise from features (can be well-generalized) instead of bugs (do not generalize due to effects of poor statistical concentration). 
The features may be robust or not robust.
Non-robust features can be easily manipulated by imperceptible noise, while robust features will not.
Still, the community needs to reach a consensus on the underlying reason for the prevalence of adversarial examples.

To further analyze the impact of adversarial examples, we propose using Information Bottleneck (IB) as a learning objective to improve adversarial robustness.
The IB is supposed to find the optimal trade-off between compression of input $X$ and prediction of $Y$~\cite{DBLP:journals/corr/Shwartz-ZivT17}.
Empirically, IB provides both performance and adversarial robustness when embedded into the learning objective.
However, computing mutual information quantity $I(X,T_l)$ is difficult in practice, especially when dealing with high-dimensional data $X$ and $T_l$.
To address this problem, Alemi et al.~\cite{DBLPconficlrAlemiFD017} proposed Variational Information Bottleneck (VIB). 
They used the internal representation of a certain intermediate layer as a stochastic encoding $Z$ of the input data $x$.
They aimed to learn the most informative representation $Z$ about the target $Y$, measured by the mutual information between their corresponding encoding values.
Their experiments also showed that VIB is robust to overfitting and adversarial attacks.
Ma et al.~\cite{Ma_Lewis_Kleijn_2020} proposed the HSIC Bottleneck and replaced mutual information with Hilbert Schmidt Independence Criterion (HSIC).
They used the HSIC bottleneck as a learning objective for every layer of the network, which is an alternative to the conventional CE loss and back-propagation.
Wang et al.~\cite{NEURIPS2021_055e31fa} proposed HBaR, which combined HSIC Bottleneck of all hidden layers and back-propagation to improve both adversarial robustness and accuracy of clean data.

Our idea is to compute mutual information (MI) between intermediate layers and their inputs or between intermediate layers and their targets. 
Then the MI is embedded into the loss function according to IB.
We summarize the following two questions about applying IB as a learning objective: 
\begin{compactenum}
    \item Which intermediate layers do we need to use?
To address question 1), we propose using only robust layers to compute MI that is then used to apply IB in the loss function.
We refer to robust layers as layers providing obviously higher accuracy than the network trained with only CE loss under PGD attack (since PGD provides good robustness against various attacks, as discussed later).
To reduce the impact of adversarial training,
we evaluate the performance of robust layers without adversarial training.
This paper empirically shows that compared to training with cross-entropy (CE) only, each layer of the network provides different degrees of robustness when applying IB as a learning objective.
Then, using robust layers for IB objective upgrades robustness to adversarial attacks.
\item Can the representation of the non-robust layers be further improved?
To address question 2), we compute a mask to remove unnecessary feature channels of convolutional layers, as the outputs of non-robust layers are extracted by subsequent convolutional layers.
When the network is trained with the loss function with IB and learns more informative features, some features are not relevant to the classification or target as they are not informative enough.
\end{compactenum}

Our evaluation consists of two parts: (1) Combining our method with state-of-the-art adversarial training methods, e.g., PGD~\cite{https://doi.org/10.48550/arxiv.1706.06083}, TRADES~\cite{pmlr-v97-zhang19p}, and MART~\cite{wang2019improving}.
Experimental results show that our method can improve the robustness of adversarial training methods.
(2) Combining without adversarial training methods.
Experimental results show that our method provides robustness compared to other IB-related techniques and plain CE.
We also find that the robustness of VGG16 mainly comes from the last convolutional block and the first two fully connected layers when using our method.
When using these three layers to compute MI, our method provides higher accuracy on adversarial examples than using all layers.
Ablation study results reveal which part of our method provides robustness and the connection among them.
In addition, applying IB as a learning objective also accelerates training convergence according to experimental results. Our implementation is available at \url{https://github.com/xiaoyunxxy/IB-RAR}.\\
Our main contributions are:
\begin{itemize}
	\item We apply IB as a regularizer to improve adversarial robustness and natural accuracy (on clean data), and we remove unnecessary features based on the regularized network.
	\item We show that our method can be naturally embedded into state-of-the-art adversarial training methods. Our method improves accuracy on adversarial examples by an average of 2.66\% against five adversarial attacks for ResNet, wide ResNet, and VGG in Tables~\ref{tab:perfvggres} and~\ref{tab:perfres18wrn}, compared to three adversarial training benchmarks.
	\item We show that our method can improve robustness for weaker methods, like plain stochastic gradient descent (SGD) trained with the CE loss function only. In the absence of adversarial training, the VGG16 network trained using our method reaches an accuracy of 35.86\% against PGD examples, while using all layers reaches 25.61\% accuracy.
\end{itemize}

\begin{figure}
\centering
\includegraphics[width = 1\linewidth]{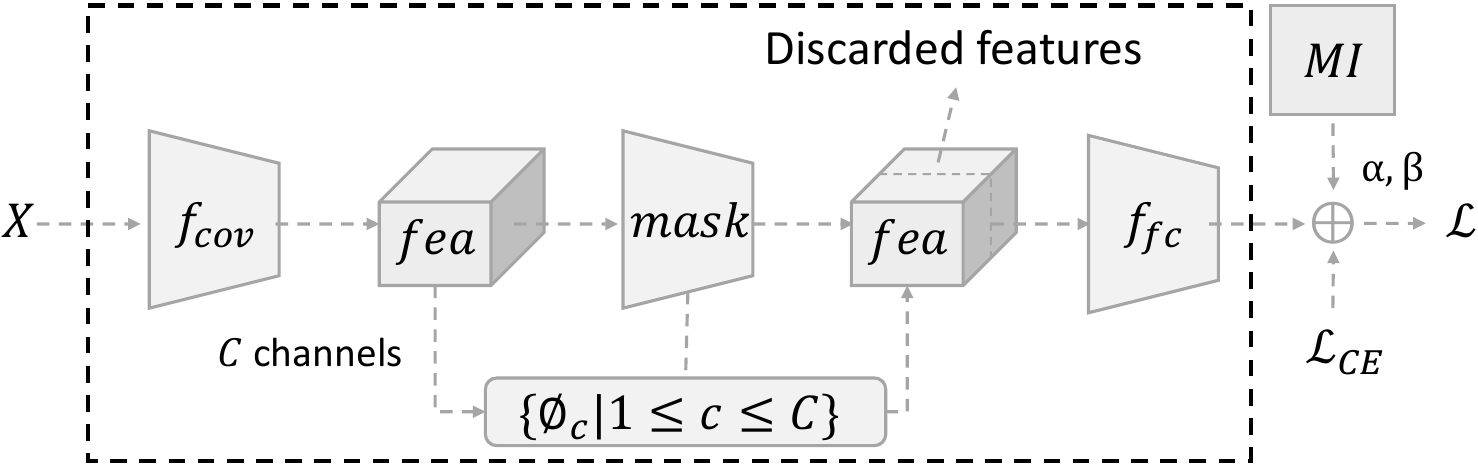}
\caption{The structure of IB-RAR. First, we use IB as a regularization combined with CE. Then, feature channels extracted from the convolutional layer are filtered by $\theta_c$.}
\label{fig:structure}
\end{figure}

\section{Methodology}
\label{methodology}

We first propose using IB as a regularizer for the learning objective when training a deep network, which also helps the network to learn better generalization of training data~\cite{DBLP:journals/corr/Shwartz-ZivT17}.
Specifically, we embed mutual information from intermediate representations to inputs $X$ and targets $Y$.
However, the outputs of convolutional layers contain independent feature channels.
We find that IB-regularized feature channels generalize better than not using IB.
Thus, we propose filtering out low correlation feature channels among well-generalized features according to their mutual information to their label.
Note that the IB regularizer is the foundation of the filtering process.
Figure~\ref{fig:structure} shows the structure of our method.

\subsection{Threat Model}
\label{threatmodel}


\textbf{Adversary Goals.}
This work focuses on adversarial attacks on image classifiers. 
The adversary aims to create imperceptible perturbation for the input image, so the network misclassifies perturbed images.
The experiments investigate untargeted individual adversarial attacks.

\textbf{Adversary Knowledge.}
Our evaluation is conducted under white-box accessibility.
The adversary has complete knowledge of the target network and its parameters.
The adversary also has full access to the training data of the target network.
The adversary is aware of the defense method, so IB-RAR is also evaluated by a specifically designed adaptive attack discussed in Section~\ref{sec:adaptiveevaluation}.

\textbf{Adversary Capabilities.}
The adversary can perturb the input to well-trained networks when the perturbation is imperceptible.
Following previous research, \textit{imperceptibility} is defined as the distance from the perturbed image to its original copy.
The distance is formalized as $l_n$-norm, i.e., $|| \delta ||_n \leq r $.

\subsection{Mutual Information Loss}
\label{methodmiloss}

\textbf{Problem Setup:}
We consider an $L$-layer neural network $F_\theta$ for classification in $d_Y$-dimensional space, 
and training dataset $D = \{ (x_i, y_i) \}_{i=0}^{n-1}$ in $d_X$-dimensional space, 
where $x_i \in \mathbb{R}^{d_X}$ and $y_i \in \{ 0,1 \}^{d_Y}$.
The $x_i$ refers to an example from training data, and we use $X$ to indicate a batch of training data in this section.
A network assigns to $x_i$ one element in $\{ 0,1 \}^{d_Y}$. 
The training aims to minimize the difference between predicted results and the ground truth, which is quantified by the standard CE loss: $\mathcal{L}_{CE}(\theta, F_\theta(x_i), y_i)$.
$T_l$ indicates the output of $l_{th}$ layer to describe the intermediate representation of a network.

The IB is embedded as a learning objective by our novel loss function:
\begin{equation}
\centering
\begin{split}
	\mathop{min}\limits_{\theta} ~ \mathcal{L} = \mathcal{L}_{CE} + \alpha \sum_{l=1}^L I(X, T_l)-\beta \sum_{l=1}^L I(Y, T_l).
\end{split}
\label{eq:miloss}
\end{equation}

Specifically, the second term $\alpha \sum_l I(X, T_l)$ minimizes the relevance between inputs and intermediate features as the loss is supposed to decrease. 
Decreasing $I(X, T_l)$ compresses input to an efficient representation, 
which naturally removes noise and irrelevant information from $X$ in $T_l$~\cite{DBLP:journals/corr/Shwartz-ZivT17}.
The term $X$ refers to a batch of inputs at each iteration while training.
The third term $\beta \sum_l I(Y, T_l)$ maximizes the relevance to the ground truth.
The term $Y$ refers to a batch of labels corresponding to $X$.
All hidden layer outputs are embedded in the loss through summations.
Because the compression measured with $I(X, T_l)$ also indicates a loss of useful information about $Y$ (as this information is indiscriminate to all content in input $X$), $I(Y, T_l)$ becomes necessary to guarantee the accuracy on clean data.
Algorithm~\ref{alg:miloss} describes how to train a network with our proposed loss from Eq.~\eqref{eq:miloss}.
The $maks$ and $T_{last}$ are discussed in Section~\ref{removeuf}.
As the computation of mutual information is difficult, we use HSIC~\cite{10.1007/11564089_7} as an alternative plan for $I(\cdot)$.

\renewcommand{\algorithmicrequire}{\textbf{Input:}}
\renewcommand{\algorithmicensure}{\textbf{Output:}}
\begin{algorithm}[t]
	\caption{Training with loss based on IB} 
	\label{alg:miloss} 
	\begin{algorithmic}
		\REQUIRE{training data $D$, network $F_\theta$ with parameters $\theta$, batch size $m$, learning rate $a$, loss $\mathcal{L}_{CE}$, optimizer SGD}
		\ENSURE{optimized weights $\theta$}
		\WHILE{Maximum epoch not reached}
		\STATE Sample $X$, a batch of data from $D$
		\STATE Forward: Calcute $F_\theta(X)$ and $T = \{ T_l | 0 \leq l < L \}$
		\STATE $T_{last} = T_{last} * mask$
		\STATE Calculate $\sum_{l=1}^L I(X, T_l)$ and $\sum_{l=1}^L I(Y, T_l)$
		\STATE Calculate loss as in Eq.~\eqref{eq:miloss}
		\STATE Backward: update $\theta$ by : $\theta \leftarrow \theta-a \nabla \mathcal{L}$ 
		\ENDWHILE 
	\end{algorithmic} 
\end{algorithm}

\textbf{Combination with adversarial training:} In addition to training with clean data, the IB loss from Eq.~\eqref{eq:miloss} can also be easily combined with adversarial training as follows:
\begin{equation}
	\begin{split}
	 \mathcal{L}_{adv}& = \mathop{max}\limits_{\delta \in S} \mathcal{L}_{CE}(\theta, F_\theta(X+\delta), X) \\
	\mathop{min}\limits_{\theta} ~ \mathcal{L}_{adv}&  + \alpha \sum_{l=1}^L I(X, T_l)-\beta \sum_{l=1}^L I(Y, T_l).
	\end{split}
	\label{eq:pgdtrain_miloss}
\end{equation}

This way, it is easy to perform adversarial training by replacing the loss in Algorithm~\ref{alg:miloss} with Eq.~\eqref{eq:pgdtrain_miloss}. 
Here, the adversarial perturbation is generated with the PGD algorithm, as it has been shown robust to various attacks~\cite{https://doi.org/10.48550/arxiv.1706.06083}.

\textbf{Selection of Robust Layers:} While other IB-related methods use one layer (such as VIB~\cite{DBLPconficlrAlemiFD017}) or all layers (such as HSIC Bottleneck~\cite{Ma_Lewis_Kleijn_2020} and HBaR~\cite{NEURIPS2021_055e31fa}), we find that deploying IB to different hidden layers will provide different robustness (see Tab.~\ref{tab:misinglelayer}).
Using a part of hidden layers for IB can get higher adversarial robustness than all layers or a single layer.
We refer to this part of layers as robust layers.
To distinguish robust layers from others, we apply IB to each hidden layer and train an independent network for each layer.
If a network shows obviously higher accuracy on adversarial examples compared to the network trained with CE only, the corresponding layer is a robust layer.

\begin{table*}[ht]
\footnotesize
\centering
\caption{Top 1 Test accuracy (in \%) on clean examples and adversarial accuracy (in \%) on adversarial examples. The adversarial examples are generated by PGD, CW, FGSM, FAB, and NIFGSM on CIFAR-10 and Tiny ImageNet. In the methods part, PGD, TRADES, and MART are benchmark methods. The method  ``(IB-RAR)'' refers to benchmarks combined with our method. Each result is the average of three runs.}
\label{tab:perfvggres}
\begin{tabular}{cccccccccccccc}
\toprule
\multicolumn{2}{l}{} & \multicolumn{6}{c}{\textbf{CIFAR-10 by VGG16}} & \multicolumn{6}{c}{\textbf{Tiny ImageNet by VGG16}} \\

\multicolumn{2}{l}{\diagbox[]{Methods}{Inputs}}	&	Natural&	PGD&	CW&	FGSM& FAB&	NIFGSM&	Natural&	PGD&	CW&	FGSM & FAB &NIFGSM\\
\midrule
\multicolumn{2}{l}{PGD}	&	75.02 & 42.45 & 37.80 & 47.32 & 41.03 & 47.59 &	37.54 & 17.73 & 13.77 & 19.46 & 13.76 & 22.14\\
\multicolumn{2}{l}{PGD (IB-RAR)}&	\textbf{76.22} & \textbf{45.09} & \textbf{41.83} & \textbf{50.53} & \textbf{46.22} & \textbf{51.93} &	\textbf{40.25} & \textbf{18.30} & \textbf{14.08} & \textbf{20.07} & \textbf{14.29} & \textbf{22.62}\\
\midrule
\multicolumn{2}{l}{TRADES}	&	73.44 & 43.92 & 38.28 & 47.94 & 41.64 & 48.41 &	36.80 & 18.13 & 13.73 & 19.57 & 14.01 & 22.16\\
\multicolumn{2}{l}{TRADES (IB-RAR)}	&	\textbf{80.63} & \textbf{44.13} & \textbf{41.81} & \textbf{51.45} & \textbf{43.63} & \textbf{51.69} & \textbf{39.10} & \textbf{18.45} & \textbf{14.19} & \textbf{20.22} & \textbf{14.49} & \textbf{22.87}\\
\midrule
\multicolumn{2}{l}{MART}	&	73.52 & 44.64 & 37.58 & 48.73 & 40.56 & 48.95 &	34.94 & 17.49 & 13.06 & 18.88 & 13.68 & 21.23\\
\multicolumn{2}{l}{MART (IB-RAR)}	&	\textbf{80.54} & 44.34 & \textbf{41.45} & \textbf{52.19} & \textbf{44.72} & \textbf{51.93} &	\textbf{36.68} & \textbf{18.05} & \textbf{13.36} & \textbf{19.33} & \textbf{13.81} & \textbf{22.02}\\
\bottomrule
\end{tabular}
\end{table*}

\begin{table*}[ht]
\footnotesize
\centering
\caption{Top 1 Test accuracy (in \%) on clean examples and adversarial accuracy (in \%) on adversarial examples. The adversarial examples are generated by PGD, CW, FGSM, FAB, and NIFGSM on CIFAR-10 and CIFAR-100. In the methods part, PGD, TRADES, and MART are benchmark methods. The ``method  (IB-RAR)'' refers to benchmarks combined with our method. Each result is the average of three runs.}
\label{tab:perfres18wrn}
\begin{tabular}{cccccccccccccc}
\toprule
\multicolumn{2}{l}{} & \multicolumn{6}{c}{\textbf{CIFAR-10 by ResNet-18}} & \multicolumn{6}{c}{\textbf{CIFAR-100 by WRN-28-10}} \\

\multicolumn{2}{l}{\diagbox[]{Methods}{Inputs}}	&	Natural&	PGD&	CW&	FGSM&	FAB&	NIFGSM&	Natural&	PGD&	CW&	FGSM&	FAB&	NIFGSM\\
\midrule
\multicolumn{2}{l}{PGD}	&	75.05 & 45.21 & 74.09 & 48.60 & 42.26 & 49.71 &	39.88 & 9.74 & 13.66 & 16.85 & 10.28 & 14.53\\
\multicolumn{2}{l}{PGD (IB-RAR)}	&	\textbf{75.10} & \textbf{45.55} & \textbf{74.10} & \textbf{48.83} & \textbf{42.74} & \textbf{50.03} &	37.68 & \textbf{16.60} & \textbf{15.98} & \textbf{19.44} & \textbf{14.85} & \textbf{19.48}\\
\midrule
\multicolumn{2}{l}{TRADES}	&	73.04 & 45.91 & 72.16 & 48.51 & 42.59 & 49.92 &	39.38 & 10.44 & 14.69 & 17.60 & 10.42 & 15.38\\
\multicolumn{2}{l}{TRADES (IB-RAR)}	&	\textbf{73.07} & \textbf{46.13} & 72.16 & \textbf{48.85} & \textbf{42.74} & \textbf{50.09} & 36.41 & \textbf{19.18} & \textbf{16.67} & \textbf{20.69} & \textbf{16.61} & \textbf{21.95}\\
\midrule
\multicolumn{2}{l}{MART}	&72.96 & 46.17 & 72.00 & 49.19 & 41.62 & 50.34 &	39.91 & 12.30 & 14.29 & 17.85 & 11.73 & 16.57\\
\multicolumn{2}{l}{MART (IB-RAR)}	&	\textbf{76.85} & \textbf{48.92} & \textbf{75.78} & \textbf{52.52} & \textbf{45.01} & \textbf{54.72} & \textbf{40.65} & \textbf{23.44} & \textbf{17.96} & \textbf{24.46} & \textbf{19.24} & \textbf{26.41}\\
\bottomrule
\end{tabular}
\end{table*}

\subsection{Removing Unnecessary Features}
\label{removeuf}
After training the network with MI loss from Eq.~\eqref{eq:miloss}, the network is supposed to learn a more informative generalization of training data.
The output of convolutional layers will have a closer connection to their targets, quantified by MI in this paper.
Specifically, feature channels containing more noise will be irrelevant to their label.
The network provides the most informative representations after extracting all convolutional layers. 
Otherwise, those convolutional layers are not necessary for the network structure.
Therefore, we evaluate feature channels given by the last convolutional layer according to their MI to the target $Y$.
In addition, there is another reason for choosing the last convolutional layer.
According to results in~\cite{NEURIPS2021_8e5e15c4}, distilled non-robust features from the last convolutional layer significantly decrease the accuracy on clean and adversarial examples.
It means non-robust (unnecessary) features have a huge negative effect on the last convolutional layer, which should be discarded.
 
Structurally, a network $F_\theta$ trained with our MI loss is given by the concatenation of $L$ hidden layers outputs $F_{\theta_{l}}$.
The network's output is the output of the last layer in the network, i.e., $F_{\theta_{L}}$.
Each layer uses the output of the previous layer as input.
We consider $F_\theta$ has $C$ kernels to extract multiple feature channels at the last convolution layer: 
$$
F_{\theta_{last}}(x) = T_{last} = \{ f_c | 1 \leq c \leq C \}.
$$

Following, we compute a mask based on the MI values of each feature channel:
\begin{equation}
\begin{aligned}
	&T_{last} = T_{last} * mask\\
	&mask = \{ \phi_c | 1 \leq c \leq C \}\\
	&\phi_c=\left\{
	\begin{aligned}
	1, ~ &I(f_c, Y) \geq thr \\
	0, ~ &otherwise.
	\end{aligned}
	\right.
\end{aligned}
\label{eq:mask}
\end{equation}

Then the mask is used to filter the feature channels.
Channels with MI less than the threshold are removed.
The threshold, $thr$, is decided according to the sorted MI values of these feature channels. 
Empirically, we use a small threshold to eliminate 5\% of all feature channels.
In other words, the MI values of that 5\% of feature channels are lower than the MI values of all other channels.
The threshold is the maximum of MI values of that 5\% of feature channels.
The application of the mask is shown in Algorithm~\ref{alg:miloss}. 
Note that removing unnecessary features is built on our MI loss, as it requires non-robust features to be more distinct from other features concerning MI values.
Experimental evidence can be found in the ablation study, row (5) of Table~\ref{tab:ablation}.

\section{Experimental Evaluation}
\label{experiments}



Following prior literature, experiments are conducted with three standard datasets,  CIFAR-10~\cite{krizhevsky2009learning}, SVHN~\cite{37648}, CIFAR-100~\cite{krizhevsky2009learning}, and Tiny ImageNet~\cite{russakovsky2015imagenet}.
We use VGG16~\cite{simonyan2014very} for CIFAR-10 and Tiny ImageNet, ResNet18~\cite{He_2016_CVPR} for CIFAR-10, and WideResNet-28-10~\cite{BMVC2016_87} for CIFAR-100.
The implementation is built with PyTorch and Torchattacks~\cite{kim2020torchattacks} frameworks.

\textbf{Algorithms}: We evaluate our method with the following adversarial learning algorithms:
Projected Gradient Descent (PGD)~\cite{https://doi.org/10.48550/arxiv.1706.06083}, TRADES~\cite{pmlr-v97-zhang19p}, and MART~\cite{wang2019improving}. 
Clean examples are not used for PGD adversarial training but are used in TRADES and MART for evaluation.
We combine our method with these algorithms and compare them against the performance of the original algorithms.
In addition, we also compare our method to non-adversarial training algorithms: Cross-Entropy, HSIC Bottleneck as Regularizer (HBaR)~\cite{NEURIPS2021_055e31fa}, and Variational Information Bottleneck (VIB)~\cite{DBLPconficlrAlemiFD017}.

\textbf{Metrics}: We evaluate accuracy on natural inputs (Test Acc., i.e., accuracy on clean data) and adversarial examples (Adv. Acc.) for all algorithms. 
The adversarial examples are generated by: (1) PGD$^n$~\cite{https://doi.org/10.48550/arxiv.1706.06083}, the PGD attack with $n$ steps in optimization; (2) FGSM~\cite{DBLP:journals/corr/GoodfellowSS14}; (3) CW~\cite{7958570}; (4) FAB~\cite{pmlr-v119-croce20a}; (5) NIFGSM~\cite{DBLP:conf/iclr/LinS00H20}. 
We set parameters for attacks (implemented by Torchattacks) following the prior literature: step size = 2/255 (alpha), r = 8/255 (eps, the limitation for perturbation $\delta$), default step s= 10, and CW steps = 200.

We use the following hyperparameters for all training: 
\begin{compactitem}
	\item StepLR: lr = 0.01, step\_size=20, gamma=0.2
	\item Optimizer: SGD, weight\_decay=1e-2
	\item Maximum epoch: 60
	\item Batch size: 100
\end{compactitem}

\textbf{Adaptive Evaluation}. To demonstrate the effectiveness of IB-RAR as a defense, we provide two levels of adaptive evaluation:
(1) To prove that the success of IB-RAR is not limited to a few cases and that the attack algorithms are converged, we use multiple attacks and iteration steps.
The results of adversarial robustness are shown in Table~\ref{tab:perfvggres}, Table~\ref{tab:perfres18wrn}, and Figure~\ref{fig:robustnesswithadv}.
(2) We assume that the adversary designs a new attack specifically targeted to IB-RAR, as the adversary has full knowledge of IB-RAR and white-box access to the network, which is discussed in Section~\ref{sec:adaptiveevaluation}.

\subsection{Adversarial Robustness Results with Adversarial Training}

We show that our method reaches better adversarial robustness along with state-of-the-art adversarial training benchmarks.
Different regularizers ($\alpha$ and $\beta$ in $\mathcal{L}$) are evaluated to find the optimal hyperparameters.
We also find that our method can boost the convergence of the network.

\subsubsection{Accuracy on Adversarial Examples}

Tables~\ref{tab:perfvggres} and~\ref{tab:perfres18wrn} show the results for test accuracy and adversarial accuracy on CIFAR-10, CIFAR-100, and Tiny ImageNet, respectively.
PGD refers to adversarial training with PGD examples.
TRADES and MART are baseline methods mentioned in the experimental setting.
The ``method (IB-RAR)'' refers to using our method to improve the baseline method, i.e., using the mutual information loss in Eq.~\eqref{eq:pgdtrain_miloss} and using the mask in Eq.~\eqref{eq:mask} to remove unnecessary feature channels.

Combined with all benchmark methods, our method improves adversarial robustness compared to baselines.
In Table~\ref{tab:perfvggres}, our method also improves the test accuracy on clean examples, especially for TRADES and MART.
Note that we use clean examples to compute MI in Eq.~\eqref{eq:pgdtrain_miloss}.
Suppose we use adversarial examples to compute MI, i.e., using $I(X+ \delta, T_l)$ to build the IB objective in the loss.
In that case, the performance increases when defending against the PGD attack (or keeping almost the same performance) but decreases the performance against other attacks.

\begin{figure*}[!ht]
	\centering
	\subfigure[PGD Attacks]{
		\begin{minipage}[b]{0.23\textwidth}
			\includegraphics[width=1\textwidth]{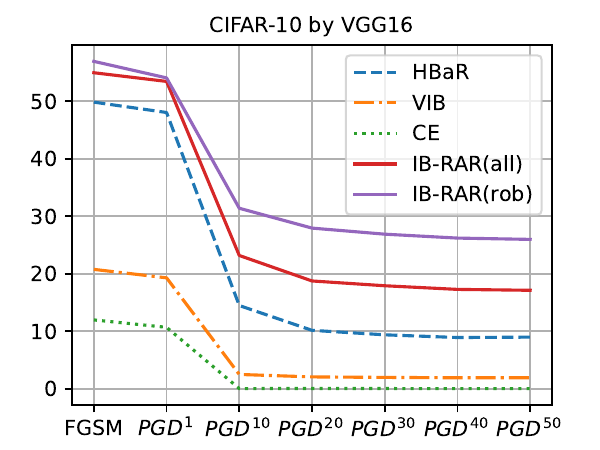} 
		\end{minipage}
		\label{fig:robustnesswithadv_a}
	}
    	\subfigure[CW Attacks]{
    		\begin{minipage}[b]{0.23\textwidth}
   		 	\includegraphics[width=1\textwidth]{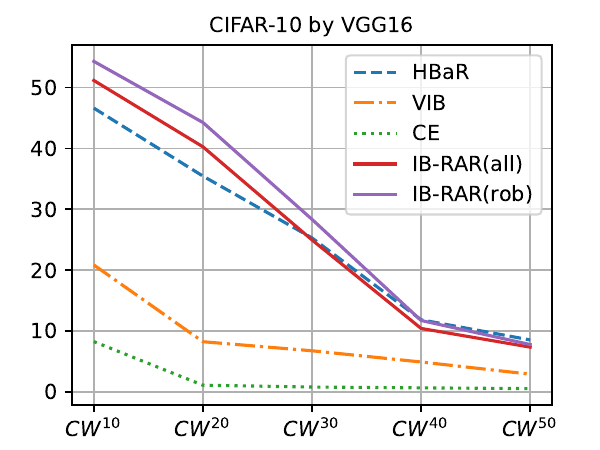}
    		\end{minipage}
		\label{fig:robustnesswithadv_b}
    	}
	\subfigure[NIFGSM Attacks]{
		\begin{minipage}[b]{0.23\textwidth}
			\includegraphics[width=1\textwidth]{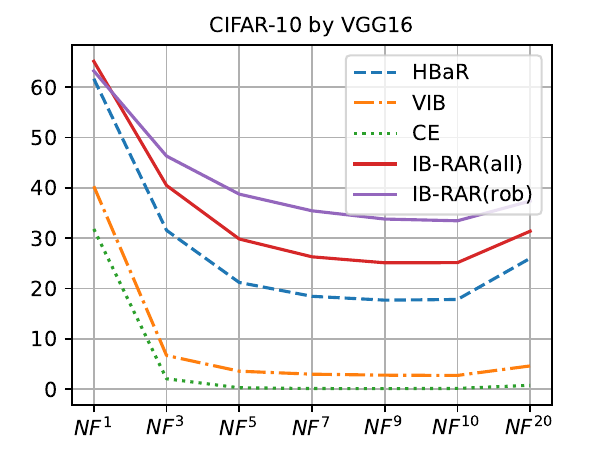} 
		\end{minipage}
		\label{fig:robustnesswithadv_c}
	}
    	\subfigure[Clean Inputs]{
    		\begin{minipage}[b]{0.23\textwidth}
		 	\includegraphics[width=1\textwidth]{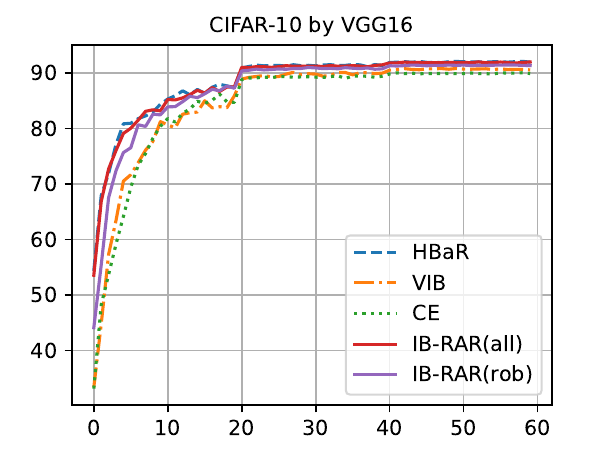}
    		\end{minipage}
		\label{fig:robustnesswithadv_d}
    	}
	\caption{CIFAR-10 with VGG16: comparison of our method and IB-based baselines. The performance is evaluated under different optimization steps of (a) PGD attacks, (b) CW attacks, (c) NIFGSM attacks, and (d) clean data. The IB-RAR(rob) refers to our method, which uses IB regularizers for only robust hidden layers. The IB-RAR(all) refers to using IB regularizers for all hidden layers. The accuracy on clean data at the last epoch is: IB-RAR(rob) 91.33\%, IB-RAR(all) 91.97\%, HBaR 91.93\%, VIB 90.52\%, CE only 89.88\%. Each result is the average of three runs.}
	\label{fig:robustnesswithadv}
\end{figure*}

\subsection{Robustness Without Adversarial Training}

\subsubsection{Using Partial Layers is Better}
\label{sec:partiallayer}
We empirically show that using partial layers to compute IB loss is better, as shown by results provided in Table~\ref{tab:misinglelayer}. 
We deploy MI loss (Eq.~\eqref{eq:miloss}) to every layer of VGG16 and use CIFAR-10 for training, as there are both convolutional and fully connected layers in the VGG structure.
Each row in Table~\ref{tab:misinglelayer} shows the performance of a network, which is trained by computing IB loss (Eq.~\eqref{eq:miloss}) for a single block of VGG16.

Clearly, the robustness mainly comes from Conv Block 5 (the fifth convolutional block in VGG16), FullyC 1 (the first fully connected layer in VGG16), and FullyC 2. 
We refer to Conv Block 5, FullyC 1, and FullyC 2 as robust layers, as they provide obvious robustness compared to other layers.
Using all layers to build the two regularizers for MI loss degrades its robustness compared to using robust layers.
Based on MI loss, we further improve its robustness on other convolutional blocks by filtering out unnecessary feature channels, row (2) and row (6) in Table~\ref{tab:ablation}.
Our method provides the best accuracy (35.86\%) under the PGD attack.
It is achieved by the defender without any prior knowledge of adversarial examples.

This phenomenon also occurs in other networks trained with other datasets, i.e., every single hidden layer can provide a different degree of adversarial robustness when computing MI of intermediate representation for IB objective. 
Their behaviors are similar but not the same.
For example, when VGG16 is trained with SVHN, the last four layers provide adversarial accuracy, i.e., Conv Block 4 (6.44\%), Conv Block 5 (16.83\%), FullyC 1 (8.97\%) and FullyC 2 (9.98\%).
When training ResNet18 with SVHN, the last layer provides higher adversarial accuracy (6.13\%) than other layers.
The accuracy of trained VGG16 and ResNet18 with CE only and SVHN dataset is lower than 1\%.
Usually, the robust layers will be the last few layers of the network.

\begin{table}[tb]
\footnotesize
\begin{center}
\begin{tabular}{ccc}
\toprule
Layer&	Adv. acc.&	Test acc.\\
\midrule
Conv Block 1&	0.04&	89.32\\
\midrule
Conv Block 2&	0.05&	90.17\\
\midrule
Conv Block 3&	0.02&	90.53\\
\midrule
Conv Block 4&	0.01&	89.66\\
\midrule
Conv Block 5&	8.25&	89.58\\
\midrule
FullyC 1&	9.85&	91.04\\
\midrule
FullyC 2&	3.27&	90.97\\
\midrule
All Layers&	25.61&	91.96\\
\midrule
Rob. Layers&	35.86&	90.97\\
\bottomrule
\end{tabular}
\end{center}
\caption{The Adv. acc and Test acc of using a single layer to compute MI in Eq.~\eqref{eq:miloss} for our method. The Adv. acc is evaluated under our default PGD attack. The network is VGG16 trained with CIFAR-10. Rob (robust) Layers refers to using outputs of Block 5, FullyC 1, and FullyC 2.}
\label{tab:misinglelayer}
\end{table}


\subsubsection{Comparison with Other IB-related Methods}
Figure~\ref{fig:robustnesswithadv} shows that our method achieves the best robustness compared to other IB-based baselines when training without adversarial examples. Specifically, we compare our method with CE, HBaR~\cite{NEURIPS2021_055e31fa}, and VIB~\cite{DBLPconficlrAlemiFD017} under the same conditions. CE refers to training with only cross-entropy loss function, i.e., no defense on this baseline.

We obtain improved accuracy on clean data compared to VIB and CE only.
Our method achieves the accuracy for IB-RAR(rob) of 91.33\%, IB-RAR(all) of 91.97\%, while HBaR, VIB, and CE only achieve 91.93\%, 90.52\%, and 89.88\%, respectively.
Results in Figure~\ref{fig:robustnesswithadv} are trained without adversarial examples.
As discussed in related works, adversarial training is recognized as the most effective defense method against adversarial attacks.
Simply using IB-based methods alone without adversarial training, including our solution, cannot perform better than the strong adversarial training on defending.
However, IB-based methods can be combined with adversarial training and improve the performance of adversarial training.
Each individual IB-based method also shows modest robustness.
However, all IB-based methods show obvious robustness compared to the CE baseline.

\subsection{Discussion and Future Work}
\label{discussion}

One possible explanation for why our method works is that there are shared features among different classes of training data.
Shared features refer to the similar characteristic of objects in two classes of data.
For example, cats and dogs are very similar, while cats and airplanes are not so similar in terms of shape. 
Clusters in Figure~\ref{fig:tsne_a} also show the similarity between classes in terms of distance.
The MI loss and mask reduce that shared feature and increase the distance among classes in Figure~\ref{fig:tsne}.

In addition, to check whether similar classes tend to be classified as each other, we evaluate the number of times the network predicted the adversarial example as a specific class (top 4 classes) 
The test set of CIFAR-10 is used to generate the adversarial examples, containing 1000 images for each class.
In Table~\ref{tab:classtendencyplain}, cars are thought to be the truck 681 times.
The highest number of classifications of the truck class is also car class, i.e., 427 times.
Such a bidirectional tendency also exists in other classes.
The network learned the most often shared features from these pairs compared to other classes.
It is easier for the adversary to find imperceptible perturbations of similar pairs, as their distance in classification should be close to each other.

This intuitive idea can be a starting point for future work that could investigate, for instance, the following aspect.
Currently, our method builds the IB objective by using inputs, outputs, and intermediate network representations.
It is not specifically designed for adversarial perturbation or shared features.
The straightforward next step is distilling shared features for every class since the shared features could help adversarial attack algorithms find small enough perturbations.
Then according to distilled features, the network can learn well-generalized features but discard shared features.
As discarding shared features may also result in the loss of useful information for the class, a key problem might be controlling the trade-off between discarding shared features and retaining enough information for generalization.

\section{Related Work}
\label{relatedwork}

The IB is supposed to find the optimal trade-off between compression of input $X$ and prediction of $Y$~\cite{DBLP:journals/corr/Shwartz-ZivT17}.
Empirically, IB provides both performance and adversarial robustness when embedded into the learning objective.
Alemi et al.~\cite{DBLPconficlrAlemiFD017} proposed Variational Information Bottleneck (VIB).
They use the internal representation of a certain intermediate layer as a stochastic encoding $Z$ of the input data $x$.
They aim to learn the most informative representation $Z$ about the target $Y$, measured by the mutual information between their corresponding encoding values.
Their experiments also showed that VIB is robust to overfitting and adversarial attacks.
Ma et al.~\cite{Ma_Lewis_Kleijn_2020} proposed the HSIC Bottleneck and replaced mutual information with Hilbert Schmidt Independence Criterion (HSIC).
They used the HSIC bottleneck as a learning objective for every layer of the network, which is an alternative to the conventional CE loss and back-propagation.
Wang et al.~\cite{NEURIPS2021_055e31fa} proposed HBaR, which combined HSIC Bottleneck of all hidden layers and back-propagation to improve both adversarial robustness and accuracy of clean data.
Previous works tend to use one layer (such as VIB) or all layers (such as HSIC Bottleneck and HBaR) for variants of IB, but we find that the IB of each layer has a different impact on robustness. 
Using partial layers will increase the robustness against adversarial attacks, as discussed in Section~\ref{sec:partiallayer}.

\section{Conclusions}
\label{conclusion}

This paper proposed an improved IB-based loss function to improve adversarial robustness and a feature channel mask to remove unnecessary features.
We first use IB as a regularizer to improve robustness on fully connected layers and learn better generalization of input data.
Based on a well-generalized network, we remove less relevant feature channels of convolutional layers according to their MI to their labels.
We also discuss that using partial layers for MI loss provides better robustness against adversarial attacks.
Our experimental results show that our method consistently improves the adversarial robustness of state-of-the-art adversarial training technologies.
Our method improves accuracy by an average of 2.66\% against five adversarial attacks for all networks in Tables~\ref{tab:perfvggres} and~\ref{tab:perfres18wrn}, compared to three adversarial training benchmarks.
Our method can also provide modest robustness to weaker methods without prior knowledge of adversarial examples.
Our findings also show that our method increases the accuracy of clean data, as the noise in input data is removed.
Finally, our experimental evaluation results demonstrate that our method can enhance the robustness against various adversarial attacks.




\bibliographystyle{plain}
\bibliography{mi_loss_ref}

\clearpage
\appendices

\section{Additional Results}

\begin{table*}[t]
\footnotesize
\begin{center}
\begin{tabular}{lcccccccc}
\toprule
 & \multicolumn{4}{c}{\textbf{CIFAR-10 with VGG16}} & \multicolumn{4}{c}{\textbf{CIFAR-10 with ResNet18}}\\
\multicolumn{1}{l}{\diagbox[]{Methods}{Inputs}} & Natural & PGD & NIFGSM & FGSM	& Natural & PGD & NIFGSM & FGSM\\
\midrule
(1) $\mathcal{L}_{CE}$&	89.99&	0.10&	0.18&	11.80&	92.19&	0.00&	0.00&	5.22 \\
\midrule
(2) $\mathcal{L}$&	92.03&	12.39&	13.90&	43.49&	93.32&	3.85&	4.71&	40.46\\
\midrule
(3) $\mathcal{L}_{CE} + \alpha \sum_{l=1}^L I(X, T_l)$&	 41.69&	0.16& 0.20& 9.98&	10.00&	10.00&	10.00&	10.00\\
\midrule
(4) $\mathcal{L}_{CE} -\beta \sum_{l=1}^L I(Y, T_l)$&	91.50&	0.06&	0.99&	31.66&	92.75&	0.00&	0.00&	8.90\\
\midrule
(5) $\mathcal{L}_{CE} + FC$&	89.41&	0.16&	0.14&	12.89&	92.41& 0.00&	0.01&	4.26\\
\midrule
(6) $\mathcal{L}+FC$ (IB-RAR)&	91.50&	35.86&	37.44& 55.92&	93.13&	5.37&	6.09&	39.34\\
\bottomrule
\end{tabular}
\end{center}
\caption{Ablation study for our method. We remove a part of our method one by one in each row. We evaluate the their natural test accuracy (in\%) and adversarial robustness (in\%) against PGD$^{10}$, NIFGSM$^{10}$, and FGSM.}
\label{tab:ablation}
\end{table*}

\subsection{Ablation Study}

\begin{figure}
\centering
\includegraphics[width = 1\linewidth]{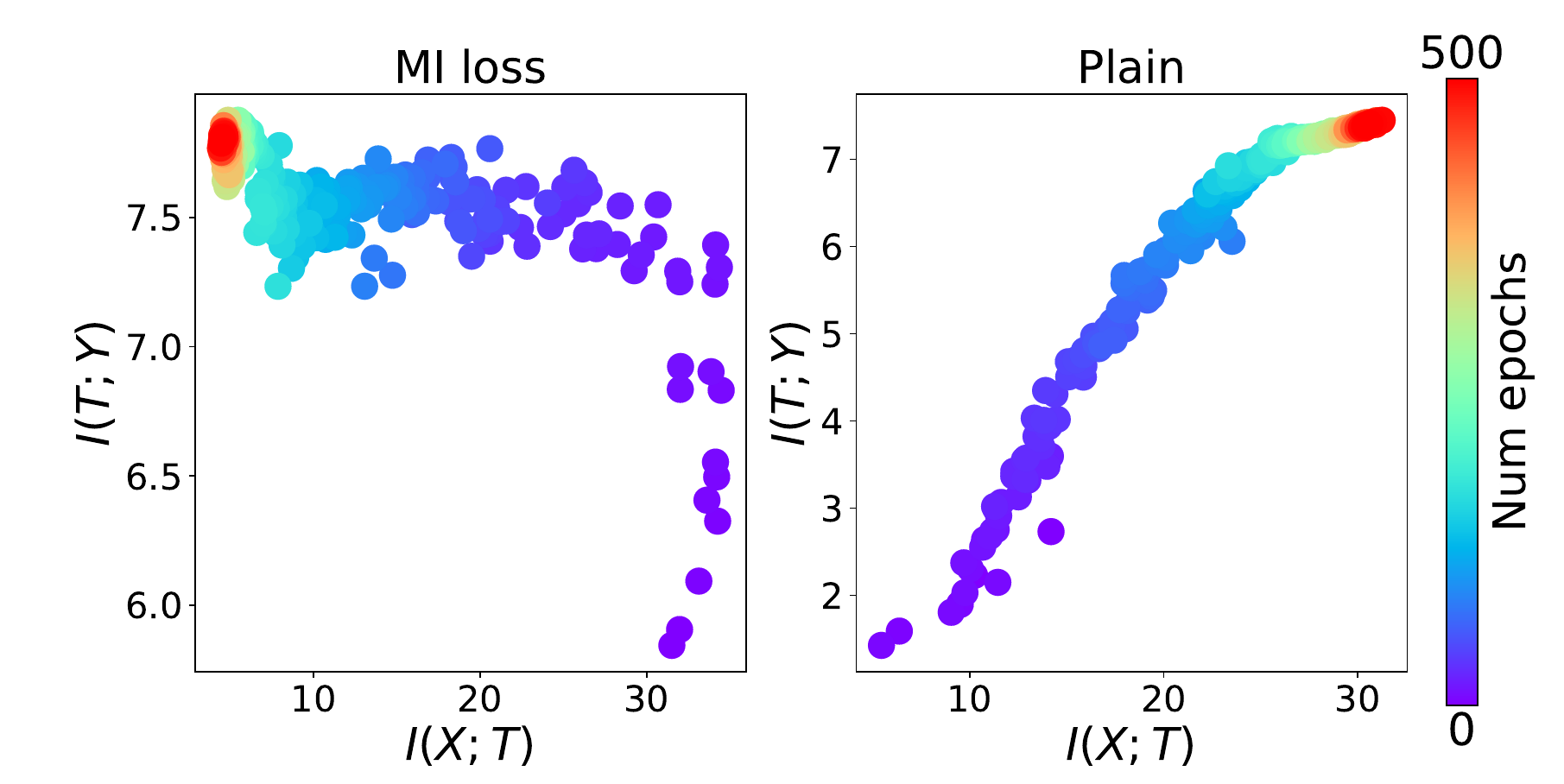}
\caption{The information plane (IP) shows the mutual information while training, recorded every 50 iterations of 60 epochs. The batch size is 100, so there are 500 iterations at every epoch. The left IP is mutual information of the 4th convolution block of VGG-16 trained with the MI loss function. The right one is the same but with only a CE loss function. The color scale indicates the training progress.}
\label{fig:vgg_cifar_mi_plain}
\end{figure}

We conduct an ablation study to verify the effectiveness of the proposed mutual information loss and the mask to remove the unnecessary feature channels.
Here, we refer to them as $\mathcal{L}$ and $FC$.
The results are shown in Table~\ref{tab:ablation}.
The network in a row (1) of Table~\ref{tab:ablation} is trained with CE loss function only.
It gets almost zero accuracies on PGD and CW and very low accuracy on FGSM, as training with CE only does not defend against adversarial attacks.
The network trained with mutual information loss, row (2) of $\mathcal{L}$, gains modest accuracy under adversarial attack, but it is lower than our method, i.e., the last row ($\mathcal(L)+FC$).
We also evaluate the regularizer terms ($I(X, T_l)$ and $I(Y, T_l)$) in mutual information loss.
Removing the enhancement of $I(Y, T_l)$, row (3), dramatically degrades the accuracy of clean data because decreasing only $I(X, T_l)$ removes both useful information and noise in inputs.
Removing the penalty of $I(X, T_l)$, row (4), gets a good network and slightly higher accuracy on clean data and adversarial examples compared to training with only CE loss in a row (1).
The reason is that increasing $I(Y, T_l)$ will increase the relevance between intermediate results (outputs of hidden layers) and their labels.
The $\mathcal{L}_{CE} + FC$ in a row (5) does not improve the robustness because the network trained with CE loss only does not learn well-generalized representation in the sense of mutual information.
In addition, as shown in Figure~\ref{fig:vgg_cifar_mi_plain}, the left figure shows a network trained with mutual information loss from Eq.~\eqref{eq:miloss}. 
The right figure shows a network trained with CE loss only.
This suggests that the network trained with CE only has no compression of $X$ and does not compress noise in $X$.

\begin{figure*}
	\centering
	\subfigure[Plain]{
		\begin{minipage}[b]{0.23\textwidth}
			\includegraphics[width=1\textwidth]{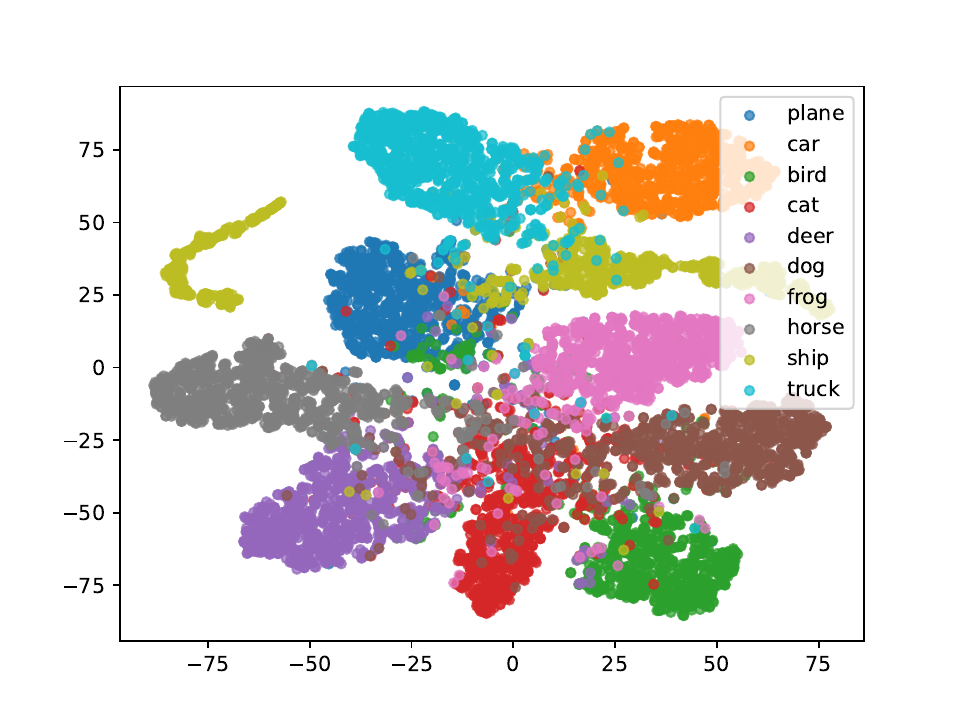} 
		\end{minipage}
		\label{fig:tsne_a}
	}
    	\subfigure[IB-RAR]{
    		\begin{minipage}[b]{0.23\textwidth}
   		 	\includegraphics[width=1\textwidth]{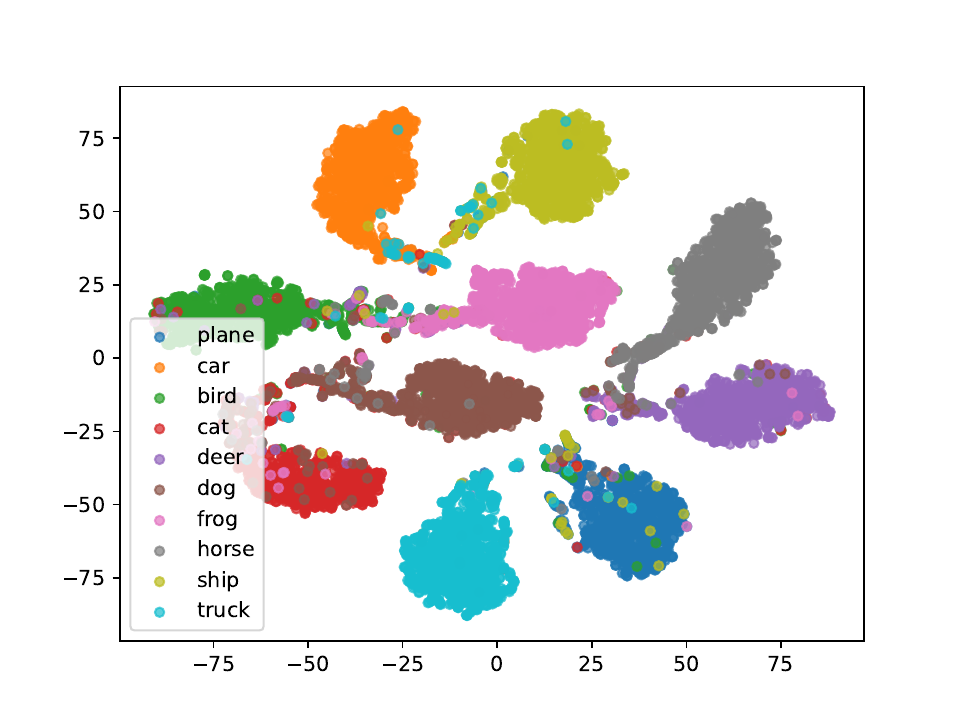}
    		\end{minipage}
		\label{fig:tsne_b}
    	}
	\subfigure[TRADES]{
		\begin{minipage}[b]{0.23\textwidth}
			\includegraphics[width=1\textwidth]{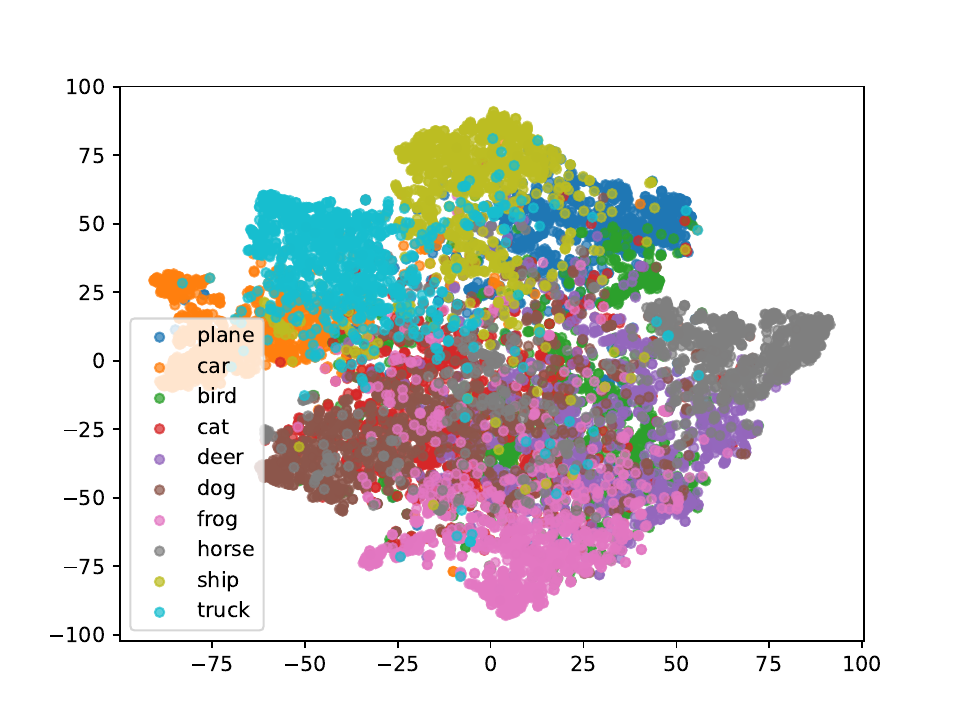} 
		\end{minipage}
		\label{fig:tsne_c}
	}
    	\subfigure[TRADES (IB-RAR)]{
    		\begin{minipage}[b]{0.23\textwidth}
		 	\includegraphics[width=1\textwidth]{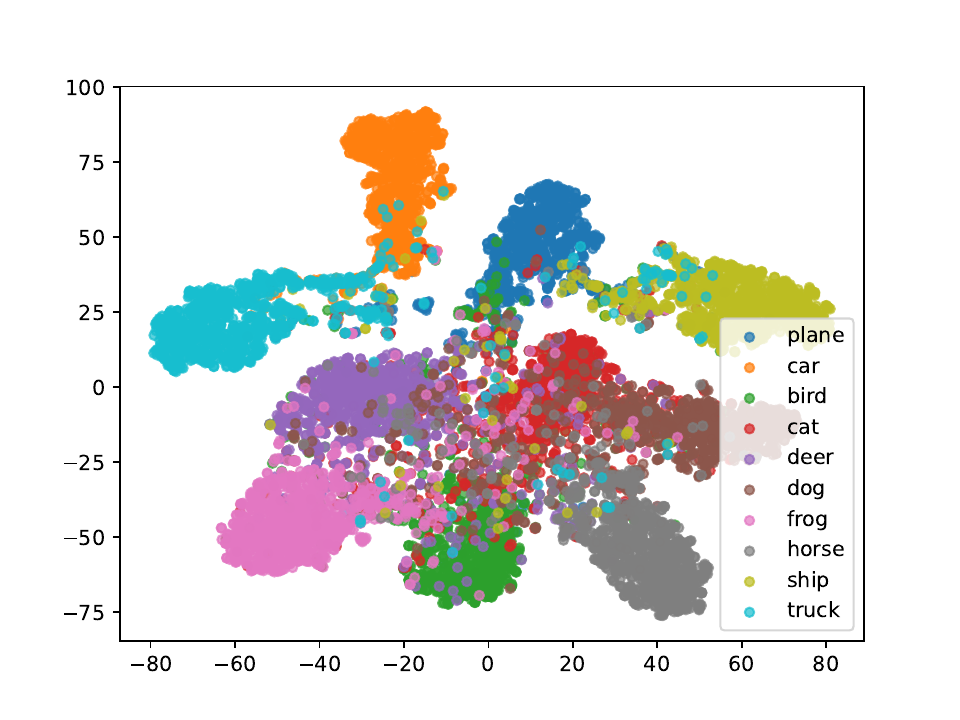}
    		\end{minipage}
		\label{fig:tsne_d}
    	}
	\caption{The results of t-SNE~\cite{JMLR:v9:vandermaaten08a} for CIFAR-10 with VGG16. Each cluster indicates a 2-dimensional feature representation from the high-dimensional representation. The feature representation is extracted from the VGG16 network. The network in (a) is trained with CE loss and clean data. The network in (b) is trained with IB-RAR and clean data. The network in (c) is trained with TRADES. The network in (d) is trained with TRADES and IB-RAR.}
	\label{fig:tsne}
\end{figure*}

To further support our experiments, we illustrate the correlation between the features generated by our method and baselines by using a 2D t-SNE plot~\cite{JMLR:v9:vandermaaten08a}.
In the case of clean examples as shown in Figures~\ref{fig:tsne_a} and~\ref{fig:tsne_b}, which are trained by CE loss only (i.e., row (1) in Table~\ref{tab:ablation}) and our method (i.e., row (6) in Table~\ref{tab:ablation}).
The accuracy (PGD) of the network in Figure~\ref{fig:tsne_b} only increases by 2.04\% (see specific values in Table~\ref{tab:ablation}) compared to the network in Figure~\ref{fig:tsne_a}.
However, it sustains better-clustered results, and a more obvious distance is maintained between clusters because the two regularizers in Eq.~\eqref{eq:miloss} remove noise in input and unnecessary features in the output of hidden layers.
Increasing the distance among clusters makes it more difficult for an adversarial attack to generate powerful perturbations,
which is why the network in Figure~\ref{fig:tsne_b} (35.86\%) reaches higher accuracy than the network in Figure~\ref{fig:tsne_a} (0.10\%) under PGD attack.
The phenomenon also happens when we deploy adversarial training methods. 
Figure~\ref{fig:tsne_c} shows interacted clusters.
Figure~\ref{fig:tsne_d} for our method shows better-clustered results compared to clusters in Figure~\ref{fig:tsne_c}.
Compared to the network in Figure~\ref{fig:tsne_c}, the natural accuracy of the network in Figure~\ref{fig:tsne_d} increases by 7.19\% (see Table~\ref{tab:perfvggres}).

This indicates that our method removes noise from inputs thanks to the $+ \alpha \sum_{l=1}^L I(X, T_l)$ regularization while keeping the relevance to the targets thanks to $-\beta \sum_{l=1}^L I(Y, T_l)$.
In addition to mutual information loss, we further filter out unnecessary features according to their mutual information with the target.
It improves robustness when adversarial attacks happen, as the adversarial noise is discarded along with unnecessary features.


\begin{table}[t]
\footnotesize
\centering
\begin{tabular}{lllll}
\toprule
\multicolumn{1}{l}{Target class} & \multicolumn{4}{c}{\textbf{Predicted results}} \\
\midrule
plane :&	bird-352  & ship-247  & deer-156  & truck-110\\
\midrule
car :&	truck-681  & ship-166  & plane-55  & frog-24\\
\midrule
bird :&	deer-260  & frog-259  & dog-141  & plane-120\\
\midrule
cat :&	dog-415  & deer-173  & bird-144  & frog-134\\
\midrule
deer :&	bird-285  & frog-196  & cat-169  & horse-147\\
\midrule
dog :&	cat-299  & frog-208  & bird-169  & horse-143\\
\midrule
frog :&	cat-411  & bird-240  & deer-187  & dog-63\\
\midrule
horse :&	dog-335  & deer-335  & truck-82  & bird-75\\
\midrule
ship :&	plane-280  & bird-196  & truck-181  & deer-116\\
\midrule
truck :&	car-427  & ship-192  & horse-135  & plane-101\\
\midrule
\end{tabular}
\caption{The adversarial example classification tendency table of CIFAR-10 trained by VGG16. The ``target class'' column is the ground truth. The rest of each row is the prediction results and the class count. Class count refers to the number of times an input of the target class (the ground truth) was classified as that class.}
\label{tab:classtendencyplain}
\end{table}


\subsection{Adaptive Attack Evaluation}
\label{sec:adaptiveevaluation}

Since IB-RAR provides a new loss function, we examine customized attacks where the adversary takes advantage of the knowledge of IB-RAR.
As the learning objective of IB-RAR is to minimize $+ \alpha \sum_{l=1}^L I(X, T_l)-\beta \sum_{l=1}^L I(Y, T_l)$ along with CE loss in Eq.~\eqref{eq:miloss} and Eq.~\eqref{eq:pgdtrain_miloss},
a natural idea is using PGD to maximize it.
We propose a white-box attack: the adversary uses $\mathcal{L}$, i.e., Eq.~\eqref{eq:miloss}, as a loss function to implement the PGD algorithm.
To ensure that the attack converges, we use 10 and 100 steps for PGD attacks.
Table~\ref{tab:adaptiveevaluation} shows the attack results.
The adaptive attack is effective because it decreases the accuracy of the network without adversarial training, i.e., plain (IB-RAR),
but the network still retains better robustness compared to training with CE only.
When combined with adversarial training, the adaptive attack does not decrease the accuracy because its robustness comes from adversarial training and IB-RAR.
If the adversary attack is specifically designed to target IB-RAR, it will weaken the attack performance of attacking adversarial training.
On the contrary, an adaptive attack targeting adversarial training will weaken the ability to attack IB-RAR.
The results demonstrate that IB-RAR is robust to the adaptive white-box attack.

\begin{table}[tb]
\footnotesize
\begin{center}
\begin{tabular}{lllll}
\toprule
Method&	PGD$_{AD}^{10}$ &PGD$^{10}$& PGD$_{AD}^{100}$ & PGD$^{100}$\\
\midrule
plain (IB-RAR)&	15.38 & 35.86 & 22.64 & 31.37\\
\midrule
AT&	45.06& 42.26 & 44.71 & 42.01\\
\midrule
AT (IB-RAR)& 45.97&	 45.03 & 45.60 & 44.60\\
\bottomrule
\end{tabular}
\end{center}
\caption{Results of IB-RAR against adaptive white-box attacks on CIFAR-10 by VGG16. PGD$_{AD}$ refers to the adaptive white box attack.}
\label{tab:adaptiveevaluation}
\end{table}

\subsection{Evaluation on Convergence}

\begin{figure}[th]
\centering
\includegraphics[width = 1\linewidth]{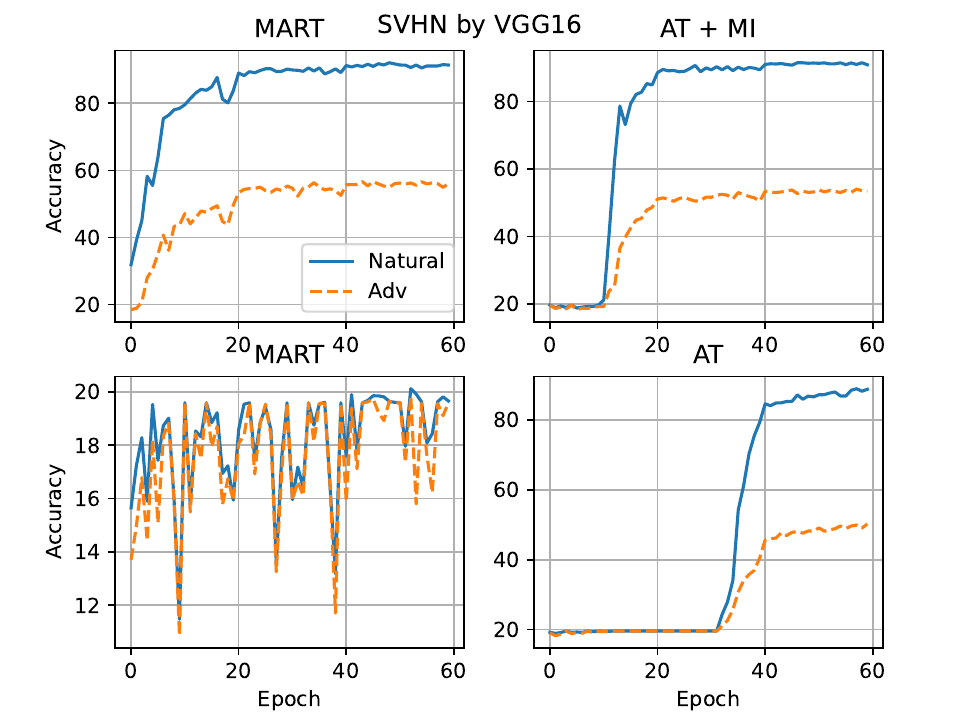}
\caption{The comparison of the accuracy of four VGG16 networks. Natural is the accuracy of clean data, and Adv is the accuracy of PGD adversarial examples. The upper-left (converged) and the lower-left (not converged) networks are trained with the SVHN dataset and MART. The converged model (upper-left) is trained with mutual information loss for only the first epoch. The upper right shows the accuracy of the network trained with the SVHN dataset, PGD adversarial training, and our method. The lower right shows the accuracy of the VGG16 network trained with the SVHN dataset, PGD adversarial training, and without our method.}
\label{fig:vgg_svhn_notconverge}
\end{figure}

Interestingly, our method also helps the network to converge faster.
We observed that the VGG16 network with SVHN does not always converge when training with MART, as shown in the lower-left part of Figure~\ref{fig:vgg_svhn_notconverge}.
When training under the same conditions, networks trained by PGD adversarial training (AT) and TRADES converge as usual.
Specifically, the network in the lower-left part of Figure~\ref{fig:vgg_svhn_notconverge} reaches the accuracy of 19.587\% during training, and then the accuracy goes lower.
After a few epochs of training, it reaches the accuracy of 19.587\% again and jumps into a loop that repeatedly reaches 19.587\%.
The network cannot break out of the loop before reaching the maximum number of epochs.
We assume that the network optimization is restricted by an under-fitting obstacle, which means the network does not capture the features of the data and results in poor accuracy.

Empirically, the problem of repeatedly reaching 19.587\% can be solved by tuning the learning rate and batch size or training for a few more epochs. 
To provide a fair comparison, we train the network with our MI loss method at the first epoch to jump out of the loop rather than tuning hyperparameters.
Then the network is trained by the CE loss function as usual.
As a result, the network converges, as shown in the top-left part of Figure~\ref{fig:vgg_svhn_notconverge}.
In addition, the top-right and lower-right of Figure~\ref{fig:vgg_svhn_notconverge} also reach the accuracy of 19.587\% at the beginning of training. Note that these networks are trained by PGD adversarial training.
The lower-right network of Figure~\ref{fig:vgg_svhn_notconverge} is trained without our method, while the top-right network is trained with our method.
After a few epochs at the beginning, both natural and adversarial accuracy of the lower-right network of Figure~\ref{fig:vgg_svhn_notconverge} reach 19.587\% and keep this value for around 30 epochs.
Then the lower-right network of Figure~\ref{fig:vgg_svhn_notconverge} increases over 19.587\%, and the natural and adversarial accuracies increase normally.
However, the top-right network of Figure~\ref{fig:vgg_svhn_notconverge} with our method breaks the under-fitting obstacle much faster.

\subsubsection{Evaluation on Different Regularizer Parameters}

Here, we evaluate our method with different regularizer hyperparameters, $\alpha$ and $\beta$ in Eq.~\eqref{eq:miloss}.
The $\alpha$ is used as $+~\alpha \sum_{l=1}^L I(X, T_l)$ and $\beta$ as $-~\beta \sum_{l=1}^L I(Y, T_l)$.
Then, the $\alpha$ parameter controls the amount of information compressed from inputs while $\beta$ controls the relevance to targets.
As shown in Table~\ref{tab:ablation}, if a network is trained with $+~\alpha \sum_{l=1}^L I(X, T_l)$ but without $-~\beta \sum_{l=1}^L I(Y, T_l)$, 
performance of the network (accuracy) is dramatically decreased.
In other words, decreasing the mutual information to inputs will remove both noise and latent useful information. 
Therefore, an $\alpha$ smaller than $\beta$ is more reasonable.
Empirically, we decide on $\alpha = \beta \times 0.1$.

\begin{figure}[tb]
	\centering
	\subfigure[Adversarial training of CIFAR-10 with VGG16]{
		\begin{minipage}[b]{0.36\textwidth}
			\includegraphics[width=1\textwidth]{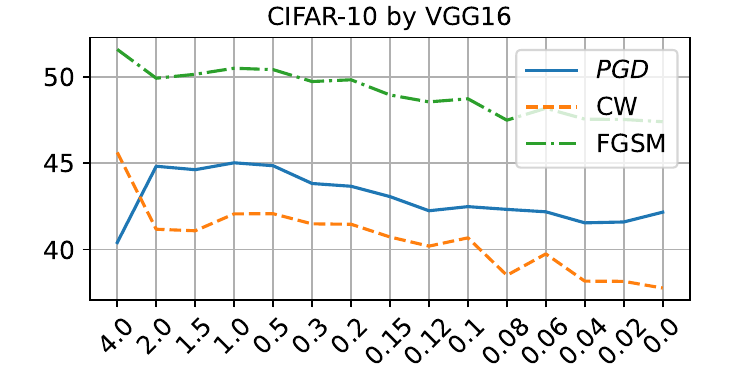} 
		\end{minipage}
		\label{fig:differentlxly_a}
	}
    	\subfigure[TRADES training of CIFAR-10 with ResNet18]{
    		\begin{minipage}[b]{0.36\textwidth}
   		 	\includegraphics[width=1\textwidth]{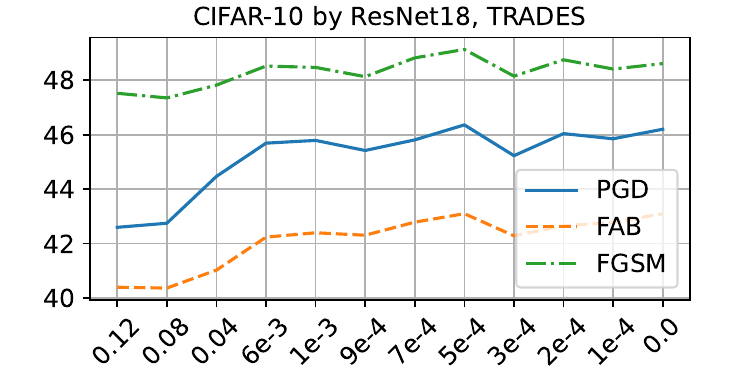}
    		\end{minipage}
		\label{fig:differentlxly_b}
    	}
	\caption{The accuracy of VGG16 and ResNet18 with different values of $\alpha$ and $\beta$ on adversarial examples. The x-axis shows values of $\beta$, $\alpha = \beta \times 0.1$. The VGG16 network is adversarially trained by PGD examples evaluated by PGD, CW, and FGSM attacks. The ResNet18 network is trained by TRADES and evaluated by PGD, FAB, and FGSM attacks.}
	\label{fig:differentlxly}
\end{figure}

The results for different values of $\alpha$ and $\beta$ are shown in Figure~\ref{fig:differentlxly}.
The goal is to select the optimal parameters for the trade-off between keeping relevant, useful information to targets and removing noise from inputs.
They are selected according to the performance of defending against the PGD attack, as the adversarial training is implemented by PGD examples.
Meanwhile, we also consider the accuracy of clean data for selecting $\alpha$ and $\beta$.
According to the results in Figure~\ref{fig:differentlxly}, we select $\alpha=1.0, \beta=0.1$ for VGG16, and $\alpha=5\times 10^{-4}, \beta=5\times 10^{-5}$ for ResNet18, which are used to obtain the results presented in Tables~\ref{tab:perfvggres} and~\ref{tab:perfres18wrn}. 
In addition, we find that the robustness is not always linearly dependent among different attacks.
Figure~\ref{fig:differentlxly_a} shows that the accuracy on PGD and the other two behave differently when $\beta$ changes from 4.0 to 2.0 and from 0.02 to 0.0.
The reason is that the adversarial training is conducted by PGD examples in the loss of Eq.~\eqref{eq:pgdtrain_miloss}.
The $+~\alpha \sum_{l=1}^L I(X+\delta, T_l)$ in Eq.~\eqref{eq:pgdtrain_miloss} decreases information about PGD examples (i.e., $X+\delta$) instead of adversarial examples of other attacks. Therefore, they behave differently in some cases.

\end{document}